\documentclass[letterpaper, 10 pt, conference]{ieeeconf}

\IEEEoverridecommandlockouts

\usepackage{amsmath}
\usepackage{amsfonts}
\usepackage{amssymb}
\usepackage{amsxtra}
\usepackage{amsthm} 
\usepackage{leftidx}
\usepackage{url}
\usepackage{graphicx,color}
\usepackage{float}
\usepackage{stfloats}
\usepackage{dsfont}
\usepackage{mathrsfs}
\usepackage{array}
\usepackage{rotating}
\usepackage{calc}
\usepackage{enumerate}
\usepackage{tikz-cd}
\usepackage{dsfont}
\usepackage{stmaryrd}
\usepackage{cancel}
\usepackage{accents}
\usepackage[linesnumbered,lined,commentsnumbered,ruled,norelsize]{algorithm2e}
\SetAlFnt{\footnotesize}
\SetAlCapFnt{\rmfamily\footnotesize}
\usepackage[colorlinks=true,urlcolor=black,linkcolor=blue,citecolor=blue]{hyperref}
\usepackage[capitalize,nameinlink]{cleveref}
\usepackage{comment}

\usepackage[normalem]{ulem}

\makeatletter
\newcommand{\removelatexerror}{\let\@latex@error\@gobble}
\makeatother

\makeatletter
\def\@endtheorem{\endtrivlist}
\makeatother

\newtheorem{theorem}{Theorem}
\newtheorem{definition}{Definition}
\newtheorem{proposition}{Proposition}
\newtheorem{remark}{Remark}

\newtheorem{lemma}{Lemma}
\newtheorem{problem}{Problem}

\date{\today}

\newcommand{\nn}{{\mathscr{N}\negthickspace\negthickspace\negthinspace\mathscr{N}}\negthinspace}
\newcommand{\ou}{%
  \mathrel{%
    \vcenter{\offinterlineskip
      \ialign{##\cr$<$\cr\noalign{\kern-1.5pt}$>$\cr}%
    }%
  }%
}%
\newcommand{\xition}[2]{\overset{#2}{_{\scriptscriptstyle\Sigma_#1} \negthickspace\negthickspace\negthinspace\negthinspace\longrightarrow}}

\newcommand{\ltsxition}[2]{\overset{#2}{_{\scriptscriptstyle #1 \vphantom{\scriptstyle (}}\negthickspace\negthickspace\negthinspace\longrightarrow}}

\newcommand{\tllTheta}{\Theta^{\overset{\text{\tiny TLL}}{~}}}

\newcommand{\subcpwa}{_{\negthinspace\text{\tiny CPWA}}}

\renewcommand{\baselinestretch}{0.91}

\begin{document}

\title{
\LARGE{\bf Two-Level Lattice Neural Network Architectures \\for Control of Nonlinear Systems
}
} %
\author{James Ferlez\textsuperscript{$*\dagger$},  Xiaowu Sun\textsuperscript{$*\dagger$}, and Yasser Shoukry\textsuperscript{$*$}
\thanks{
\textsuperscript{$\dagger$} Equally contributing first authors.
}
\thanks{
\textsuperscript{$*$}Department of Electrical Engineering and Computer Science, University of California, Irvine
\texttt{\{jferlez,xiaowus,yshoukry\}@uci.edu}
} %
\thanks{This  work  was  partially  sponsored  by  the  NSF  awards \#CNS-2002405 and \#CNS-2013824.}
} %

\maketitle

\begin{abstract}
	In this paper, we consider the problem of automatically designing a 
	Rectified Linear Unit (ReLU) Neural Network (NN) architecture (number of 
	layers and number of neurons per layer) with the guarantee that it is 
	sufficiently parametrized to control a nonlinear system. Whereas current 
	state-of-the-art techniques are based on hand-picked architectures or 
	heuristic-based search to find such NN architectures, our approach exploits 
	a given model of the system to design an architecture; as a result, we 
	provide a guarantee that the resulting NN architecture is sufficient to 
	implement a controller that satisfies an achievable specification. Our 
	approach exploits two basic ideas. First, we assume that the system can be 
	controlled by a Lipschitz-continuous state-feedback controller that is 
	unknown but whose Lipschitz constant is upper-bounded by a known constant; 
	then using this assumption, we bound the number of affine functions needed 
	to construct a Continuous Piecewise Affine (CPWA) function that can 
	approximate the unknown Lipschitz-continuous controller. Second, we utilize 
	the authors' recent results on  the Two-Level Lattice (TLL) NN 
	architecture, a novel NN architecture that was shown to be parameterized 
	directly by the number of affine functions that comprise the CPWA function 
	it realizes. 
%
%
	We also evaluate our method by designing a NN architecture to control an 
	inverted pendulum.
\end{abstract}

\includecomment{proofs} %
\excludecomment{techreport} %
\excludecomment{arxivreference}



\section{Introduction} 
\label{sec:introduction}

Multilayer Neural Networks (NN) have shown tremendous success in realizing feedback controllers that can achieve several complex control tasks~\cite{bojarski2016end}.
Nevertheless, the current state-of-the-art practices for designing these deep 
NN-based controllers are based on heuristics and hand-picked hyper-parameters 
(e.g., number of layers, number of neurons per layer, training parameters, 
training algorithm) without an underlying theory that guides their design. For 
example, several researchers have studied the problem of Automatic Machine 
Learning (AutoML) and in particular the problem of hyperparameter (number of 
layers, number of neurons per layer, and learning algorithm parameters) 
optimization and tuning for deep NNs (see for 
example~\cite{pedregosa2016hyperparameter, bergstra2012random, paul2019fast, 
baker2016designing, quanming2018taking} and the references within). 
These methods perform an iterative and exhaustive search through a manually 
specified subset of the hyperparameter space; 
the best hyperparameters are then selected according to some performance metric 
without any guarantee on the correctness of the chosen architecture.

In this paper, we exhibit a systematic methodology for choosing a NN controller 
architecture (number of layers and number of neurons per layer) to control a 
nonlinear system. Specifically, we design an architecture that is guaranteed to 
correctly control a nonlinear system in the following sense: there exist neuron 
weights/biases for the designed architecture such that it can meet exactly the 
same specification as any other continuous, non-NN controller with at most an a 
priori specified Lipschitz constant. Moreover, provided such a non-NN 
controller exists (to help ensure well-posedness), the design of our 
architecture requires only knowledge of a bound on such a controller's 
Lipschitz constant; the robustness of the specification; and the Lipschitz 
constants/vector field bound of the nonlinear system. Thus, our approach may be 
applicable even without perfect knowledge of the underlying system dynamics, 
albeit at the expense of designing rather larger architectures.




Our approach exploits several insights. First, state-of-the-art NNs use 
Rectified Linear Units (ReLU), which in turn restricts such NN controllers to 
implement only Continuous Piecewise Affine (CPWA) functions. As is widely 
known, a CPWA function is compromised of several affine functions (named local 
linear functions), which are defined over a set of polytypic regions (called 
local linear regions). In other words, a ReLU NN---by virtue of its CPWA 
character---partitions its input space into a set of polytypic regions (named 
activation regions), and applies a linear controller at each of these regions. 
Therefore, a NN architecture dictates the number of such activation regions in 
the corresponding CPWA function that is represented by the trainable parameters 
in the NN. That is, to design a NN architecture, one needs to perform two 
steps: (i) compute (or upper bound) the number of activation regions required 
to implement a controller that satisfies the specifications; and (ii) transform 
this number of activation regions into a NN architecture that is guaranteed to 
give rise to this number of activation regions.

To count the number of the required activation regions, we start by assuming 
the existence of a Lipschitz-continuous, state-feedback controller that can 
robustly control the nonlinear system to meet given specifications. 
However, as stated above, we make no further assumptions about this controller 
except that its Lipschitz constant is upper-bounded by a known constant, 
$K_\text{cont}$. Using this Lipschitz-constant bound -- \emph{but no other 
specific information about the controller} -- together with the Lipschitz 
constants/vector field bound of the system and robustness of the 
specification, we exhibit an upper-bound for the number of activation regions 
needed to approximate this controller by a CPWA controller, while still meeting 
the same specifications in closed loop. 

Next, we leverage this bound on activation regions using the authors' recent 
results on a novel NN architecture, the Two-Level Lattice (TLL) NN 
architecture~\cite{FerlezAReNAssuredReLU2020}. Unlike other NN architectures 
where the number of activation regions is not explicitly specified, the TLL-NN 
architecture is explicitly parametrized by the number of activation regions it 
contains. Thus, we can directly specify a TLL architecture from the 
aforementioned bound on the number of activation regions. 
The resulting NN architecture is then guaranteed to be sufficiently 
parametrized to implement a CPWA function that approximates the unknown 
Lipschitz-continuous controller in such a way that the specification is still 
met. This provides a systematic approach to designing a NN architecture for 
such systems.

%


\section{Preliminaries} 
\label{sec:prelims}

\subsection{Notation} 
\label{sub:notation}

We denote by $\mathbb{N}$, $\mathbb{R}$ and $\mathbb{R}^+$ the set of natural 
numbers, the set of real numbers and the set of non-negative real numbers, 
respectively. For a function $f : A \rightarrow B$, let $\text{dom}(f)$ return 
the domain of $f$, and let $\text{range}(f)$ return the range of $f$. For a set 
$V \subseteq \mathbb{R}^n$, let $\text{int}(V)$ return the interior of $V$. For 
$x \in \mathbb{R}^n$, we will denote by $\lVert x \rVert$ the infinity norm of 
$x$; for $x \in \mathbb{R}^n$ and $\epsilon \geq 0$ we will denote by 
$B(x;\epsilon)$ the ball of radius $\epsilon$ centered at $x$ as specified by 
$\lVert \cdot \rVert$. For $f : \mathbb{R}^n \rightarrow \mathbb{R}^m$, $\lVert 
f \rVert_\infty$ will denote the essential supremum norm of $f$. Finally, given 
two sets $A$ and $B$ denote by $B^A$ the set of all functions $f: A \rightarrow 
B$.


\subsection{Dynamical Model} 
\label{sub:dynamical_model}
In this paper, we will assume an underlying, but not necessarily known,  
continuous-time nonlinear dynamical system specified by an ordinary 
differential equation (ODE): that is
\begin{equation}\label{eq:main_ode}
	\dot{x}(t) = f( x(t) , u(t))
\end{equation}
where the state vector $x(t) \in \mathbb{R}^n$ and the control vector $u(t) \in 
\mathbb{R}^m$. Formally, we have the following definition:
\begin{definition}[Control System]
\label{def:control_system}
	A \textbf{control system} is a tuple $\Sigma = (X, U, \mathcal{U}, f)$ 
	where

	\begin{itemize}
		\item $X \subset \mathbb{R}^n$ is the compact subset of the state 
			space;

		\item $U \subset \mathbb{R}^m$ is the compact set of admissible 
			(instantaneous) controls;

		\item $\mathcal{U} \subseteq U^{\mathbb{R}^+}$ is the space of 
			admissible open-loop control functions -- i.e. $v \in \mathcal{U}$ 
			is a function $v : \mathbb{R}^+ \rightarrow U$; and

		\item $f : \mathbb{R}^n \times U \rightarrow \mathbb{R}^n$ is a 
			vector field specifying the time evolution of states according to 
			\eqref{eq:main_ode}.
	\end{itemize}
	A control system is said to be (globally) \textbf{Lipschitz} if there 
	exists constants $K_x$ and $K_u$ such that for all $x,x^\prime 
	\in\mathbb{R}^n$ and $u,u^\prime \in \mathbb{R}^m$: 
	\begin{equation}\label{eq:dynamics_lipshitz}
		\lVert f(x,u) - f(x^\prime, u^\prime) \rVert \leq 
			K_x \lVert x - x^\prime \rVert + K_u \lVert u - u^\prime \rVert.
	\end{equation}
	For a Lipschitz control system, the following vector field bound is well 
	defined:
	\begin{equation}
		\mathcal{K} \triangleq \max_{x\in X, u \in U} \lVert f(x,u) \rVert.
	\end{equation}
\end{definition}

In the sequel, we will primarily be concerned with solutions to 
\eqref{eq:main_ode} that result from instantaneous state-feedback controllers, 
$\Psi : X \rightarrow U$. Thus, we use $\zeta_{x_0 \Psi}$ to denote the 
\emph{closed-loop} solution of \eqref{eq:main_ode} starting from initial 
condition $x_0$ (at time $t=0$) and using \emph{state-feedback controller} 
$\Psi$. We refer to such a $\zeta_{x_0\Psi}$ as a (closed-loop) 
\emph{trajectory} of its associated control system.

\begin{definition}[Closed-loop Trajectory]
	Let $\Sigma$ be a Lipschitz control system, and let $\Psi : \mathbb{R}^n 
	\rightarrow U$ be a globally Lipschitz continuous function. A 
	\textbf{closed-loop trajectory} of $\Sigma$ under controller $\Psi$ and  
	starting from $x_0 \in X$ is the function $\zeta_{x_0\Psi} : \mathbb{R}^+ 
	\rightarrow X$ that uniquely solves the integral equation:
	\begin{equation}\label{eq:feedback_integral_eq}
		\zeta_{x_0\Psi}(t) = x_0 + \int_0^t f( \zeta_{x_0\Psi}(\sigma), \Psi( \zeta_{x_0\Psi}(\sigma) )) d\sigma.
	\end{equation}
	It is well known that such solutions exist and are unique under these 
	assumptions \cite{KhalilNonlinearSystems2001}.
\end{definition}
\begin{definition}[Feedback Controllable]
	A Lipschitz control system $\Sigma$ is \textbf{feedback controllable} by a 
	Lipschitz controller $\Psi: \mathbb{R}^n \rightarrow U$ if the following is 
	satisfied:
	\begin{equation}
		\Psi \circ \zeta_{x\Psi} \in \mathcal{U} \qquad \forall x\in X.
	\end{equation} 
	If $\Sigma$ is feedback controllable for any such $\Psi$, then we simply 
	say that it is feedback controllable.
\end{definition}
Because we're interested in a compact set of states, $X$, we consider only 
feedback controllers whose closed-loop trajectories stay within $X$.
\begin{definition}[Positive Invariance]
	A feedback trajectory of a Lipschitz control system, $\zeta_{x_0\Psi}$, is 
	\textbf{positively invariant} if $\zeta_{x_0\Psi}(t) \in X$ for all $t \geq 
	0$. A controller $\Psi$ is positively invariant if $\zeta_{x_0\Psi}$ is 
	positively invariant for all $x_0 \in X$.
\end{definition}
For technical reasons, we will also need the following stronger notion of 
positive invariance.
\begin{definition}[$\delta$,$\tau$ Positive Invariance]
\label{def:delta_tau_positive_invariance}
	Let $\delta,\tau \negthinspace > \negthinspace 0$ and 
	$\text{edge}_\delta(X) \negthinspace \triangleq \negthinspace \cup_{x \in X 
	\backslash \text{int}(X)} ( X \negthinspace \cap \negthinspace B(x;\delta) 
	)$. Then a positively invariant controller $\negthinspace \Psi 
	\negthinspace$ is $\mathbf{\delta}$,$\mathbf{\tau}$ \textbf{positively 
	invariant} if \vspace{-0.2em}
	\begin{equation}
		\forall x_0 \in \text{edge}_{\delta}(X)
		\; . \; 
		\zeta_{x_0\negthinspace\Psi}(\tau) \negthinspace \in \negthinspace X \backslash \text{edge}_{\delta}(X)
	\end{equation}
	\vspace{-14pt}

	\noindent and $\Psi$ is positively invariant with respect to $X \backslash 
	\text{edge}_\delta(X)$.
\end{definition}
For a $\delta$,$\tau$ positively invariant controller, trajectories that start 
$\delta$-close to the boundary of $X$ end up at least $\delta$-far \emph{away} 
from that boundary after $\tau$ seconds, and remain there forever after.



Finally, borrowing from \cite{ZamaniSymbolicModelsNonlinear2012}, we define a 
$\tau$-sampled transition system embedding of a feedback-controlled system.
\begin{definition}[$\tau$-sampled Transition System Embedding]
\label{def:embedding}
	Let $\Sigma=(X,U,\mathcal{U},f)$ be a feedback controllable Lipschitz 
	control system, and let $\Psi : \mathbb{R}^n \rightarrow U$ be a Lipschitz 
	continuous feedback controller. For any $\tau > 0$, the 
	$\mathbf{\tau}$\textbf{-sampled transition system embedding} of $\Sigma$ 
	under $\Psi$ is the tuple $S_\tau(\Sigma_\Psi) = (X_\tau, \mathcal{U}_\tau, 
	\xition{\Psi}{~} )$ where:

	\begin{itemize}
		\item $X_\tau = X$ is the state space;

		\item $\mathcal{U}_\tau = \{ (\Psi \circ \zeta_{x_0\Psi})|_{t\in 
			[0,\tau]}: x_0 \in X \}$ is the set of open loop control inputs 
			generated by $\Psi$-feedback, each restricted to the domain 
			$[0,\tau]$; and

		\item $\xition{\Psi}{~} \subseteq X_\tau \times \mathcal{U}_\tau 
			\times X_\tau$ such that $x \xition{\Psi}{u} x^\prime$ iff \\ 
			$\qquad$ both $u = (\Psi \circ \zeta_{x\Psi})|_{t\in [0,\tau]}$ and 
			$x^\prime = \zeta_{x\Psi}(\tau)$. 
	\end{itemize}
	$S_\tau(\Sigma_\Psi)$ is thus a \textbf{metric} transition system 
	\cite{ZamaniSymbolicModelsNonlinear2012}.
\end{definition}

\subsection{Abstract Disturbance Simulation} 
\label{sub:perturbation_}
In this subsection, we propose a new simulation relation, which we call 
\emph{abstract disturbance simulation}, as a formal notion of specification 
satisfaction for metric transition systems. 
%
Abstract disturbance simulation is inspired by robust bisimulation 
\cite{KurtzRobustApproximateSimulation2020} and especially  disturbance 
bisimulation \cite{MallikCompositionalSynthesisFiniteState2019}, but it 
abstracts those notions away from their definitions in terms of control system 
embeddings and explicit modeling of disturbance inputs. In this way, it is 
conceptually similar to the technique used in 
\cite{ZamaniSymbolicModelsNonlinear2012} and 
\cite{PolaApproximatelyBisimilarSymbolic2008} to define a quantized 
abstraction, where deliberate non-determinism is introduced in order to account 
for input errors.
As a prerequisite, we introduce the following definition.
\begin{definition}[Perturbed Metric Transition System]
\label{def:perturbed_system}
	Let $S = (X, U, \ltsxition{S}{~})$ be a metric transition system where $X 
	\subseteq X_M$ for a metric space $(X_M, d)$. Then the 
	\textbf{$\delta$-perturbed metric transition system} of $S$, 
	$\mathfrak{S}^\delta$, is a tuple $\mathfrak{S}^\delta = (X, U, 
	\ltsxition{\mathfrak{S}\negthinspace\overset{\delta}{~} 
	\thickspace\thinspace}{~})$ where the (altered) transition relation, 
	$\ltsxition{\mathfrak{S}\negthinspace\overset{\delta}{~} 
	\thickspace\thinspace}{~}$, is defined as:
	\begin{multline}
		x \thickspace 
		\ltsxition{\mathfrak{S}\negthinspace\overset{\delta}{~} \thickspace\thinspace}{u}
		x^\prime
		\text{ iff } \\
		\exists x^{\prime\prime} \in X \text{ s.t. } d(x^{\prime\prime},x^\prime) \leq \delta \text{ and } x \ltsxition{S}{u} x^{\prime\prime}.
	\end{multline}
\end{definition}
Note that $\mathfrak{S}^\delta$ has identical states and input labels to $S$, 
and it also subsumes all of the transitions therein, i.e. $\ltsxition{S}{~} 
\thinspace \subset \thinspace 
\ltsxition{\mathfrak{S}\negthinspace\overset{\delta}{~} 
\thickspace\thinspace}{~}$. However, the transition relation for 
$\mathfrak{S}^\delta$ explicitly contains new nondeterminism relative to the 
transition relation of $S$. This nondeterminism can be thought of as perturbing 
the target state of each transition in $S$; each such perturbation becomes the 
target of a (nondeterministic) transition with the same input label as the 
original transition.

Using this definition, we can finally define an 
abstract disturbance simulation between two metric transition systems.
\begin{definition}[Abstract Disturbance Simulation]
\label{def:delta_perturbation_simulation} 
	Let $S = (X_S, U, \ltsxition{S}{~})$ and $T = (X_T, U_T, \ltsxition{T}{~})$ 
	be metric transition systems whose state spaces $X_S$ and $X_T$ are subsets 
	of the same metric space $(X_M, d)$. Then $T$ \textbf{abstract-disturbance 
	simulates} $S$ under disturbance $\delta$, written $S 
	\preceq_{{\mathcal{AD}_\delta}} T$ if there is a relation $R \subseteq X_S 
	\times X_T$ such that

	\begin{enumerate}
		\item for every $(x,y) \in R$, $d(x,y) \leq \delta$;

		\item for every $x \in X_S$ there exists a pair $(x,y) \in R$; and

		\item for every $(x,y) \in R$ and $x 
			\ltsxition{\mathfrak{S}\negthinspace\overset{\delta}{~} 
			\thickspace\thinspace}{u} x^\prime$ there exists a $y 
			\ltsxition{T}{v} y^\prime$ such that $(x^\prime, y^\prime) \in R$.
	\end{enumerate}
\end{definition}

\begin{remark}
	$\preceq_{\mathcal{AD}_0}$ corresponds with the usual notion of simulation 
	for metric transition systems. Thus,
	\begin{equation}
		S \preceq_{\mathcal{AD}_\delta} T \Leftrightarrow \mathfrak{S}^\delta \preceq_{\mathcal{AD}_0} T.
	\end{equation}
\end{remark}


\subsection{ReLU Neural Network Architectures} 
\label{sub:relu_neural_network_architectures}
We will consider controlling the nonlinear system defined in 
\eqref{eq:main_ode} with a state-feedback neural network controller $\nn$:
\begin{equation}
	\nn: X \rightarrow U
\end{equation}
where $\nn$ denotes a Rectified Linear Unit Neural Network (ReLU NN).
Such a ($K$-layer) ReLU NN is specified by composing $K$ \emph{layer} functions 
(or just \emph{layers}). A layer with $\mathfrak{i}$ inputs and $\mathfrak{o}$ 
outputs is specified by a $(\mathfrak{o} \times \mathfrak{i} )$ matrix of 
\emph{weights}, $W$, and a $(\mathfrak{o} \times 1)$ matrix of \emph{biases}, 
$b$, as follows:
\begin{align}
	L_{\theta} : \mathbb{R}^{\mathfrak{i}} &\rightarrow \mathbb{R}^{\mathfrak{o}} \notag\\
	      z &\mapsto \max\{ W z + b, 0 \}
\end{align}
where the $\max$ function is taken element-wise, and $\theta \triangleq (W,b)$ 
for brevity. Thus, a $K$-layer ReLU NN function 
is specified by $K$ 
layer functions $\{L_{\theta^{(i)}} : i = 1, \dots, K\}$ whose input and output 
dimensions are \emph{composable}: that is they satisfy $\mathfrak{i}_{i} = 
\mathfrak{o}_{i-1}: i = 2, \dots, K$. Specifically:
\begin{equation}
	\nn(x) = (L_{\theta^{(K)}} \circ L_{\theta^{(K-1)}} \circ \dots \circ L_{\theta^{(1)}})(x).
\end{equation}
When we wish to make the dependence on parameters explicit, we will index a 
ReLU function $\nn$ by a \emph{list of matrices} $\Theta \triangleq ( 
\theta^{(1)}, \dots , \theta^{(K)} )$ \footnote{That is $\Theta$ is not the 
concatenation of the $\theta^{(i)}$ into a single large matrix, so it preserves 
information about the sizes of the constituent $\theta^{(i)}$.}.

Specifying the number of layers and the \emph{dimensions} of the associated 
matrices $\theta^{(i)} = (\; W^{(i)}, b^{(i)}\; )$ specifies the 
\emph{architecture} of the ReLU NN. Therefore, we will use:
\begin{equation}
	\text{Arch}(\Theta) \negthinspace \triangleq \negthinspace ( (n,\mathfrak{o}_{1}), (\mathfrak{i}_{2},\mathfrak{o}_{2}), \ldots, 
	(\mathfrak{i}_{K}, m))
\end{equation}
to denote the architecture of the ReLU NN $\nn_{\Theta}$.

Since we are interested in designing ReLU architectures, we will also need the 
following result from \cite[Theorem 7]{FerlezAReNAssuredReLU2020}, which states 
that a Continuous, Piecewise Affine (CPWA) function, $\mathsf{f}$, can be 
implemented exactly using a Two-Level-Lattice (TLL) NN architecture that is 
parameterized exclusively by the number of local linear functions in 
$\mathfrak{f}$.

\begin{definition}[Local Linear Function]
	Let $\mathsf{f} : \mathbb{R}^n \rightarrow \mathbb{R}^m$ be CPWA. Then a 
	\textbf{local linear function of} $\mathsf{f}$ is a linear function $\ell : 
	\mathbb{R}^n \rightarrow \mathbb{R}^m$ if there exists an open set 
	$\mathfrak{O}$ such that $\ell(x) = \mathsf{f}(x)$ for all $x\in 
	\mathfrak{O}$.
\end{definition}

\begin{definition}[Linear Region]
	Let $\mathsf{f} : \mathbb{R}^n \rightarrow \mathbb{R}^m$ be CPWA. Then a 
	\textbf{linear region of} $\mathsf{f}$ is the largest set $\mathfrak{R} 
	\subseteq \mathbb{R}^n$ such that $\mathsf{f}$ has only one local linear 
	function on $\text{int}(\mathfrak{R})$.
\end{definition}

\begin{theorem}[Two-Level-Lattice (TLL) NN Architecture {[}7, Theorem 7{]}]
\label{thm:tll_architecture}
	Let $\mathsf{f}:\mathbb{R}^n \rightarrow \mathbb{R}^m$ be a CPWA function, 
	and let $\bar{N}$ be an upper bound on the number of local linear functions 
	in $\mathsf{f}$. Then there is a Two-Level-Lattice (TLL) NN architecture 
	$\text{Arch}(\tllTheta_{\bar{N}})$ parameterized by $\bar{N}$ and values of 
	$\tllTheta_{\bar{N}}$ such that:
	\begin{equation}
		\mathsf{f}(x) = \nn_{\negthickspace\tllTheta_{\bar{N}}}(x).
	\end{equation}
	In particular, the number of linear regions of $\mathsf{f}$ is such an 
	upper bound on the number of local linear functions.
\end{theorem}

Finally, note that a ReLU NN function, $\nn$, is known to be a continuous, 
piecewise affine (CPWA) function consisting of finitely many linear segments. 
Thus, $\nn$ is itself necessarily globally Lipschitz continuous.



\section{Problem Formulation} 
\label{sec:problem_formulation}

We can now state the main problem that we will consider in this paper. In 
brief, we wish to identify the architecture for a ReLU network to be used as a 
feedback controller for the control system $\Sigma$: this architecture must 
have parameter weights that allow it to control $\Sigma$ up to a specification 
that can be met by some other, non-NN controller.


Despite our choice to consider fundamentally continuous-time models, we 
formulate our main problem in terms of their ($\tau$-sampled) transition system 
embeddings. This choice reflects recent success in verifying specifications for 
such transition system embeddings by means of techniques adapted from computer 
science; see e.g. \cite{TabuadaVerificationControlHybrid2009}, where a variety 
of specifications are considered in this context, among them LTL formula 
satisfaction. Thus, our main problem is stated in terms of the simulation 
relations in the previous section.

\begin{problem}
\label{prob:main_problem_generic}
	Let $\delta \negthinspace > \negthinspace 0$ and $K_\text{cont} 
	\negthinspace > \negthinspace 0$ be given. Let $\Sigma$ be a feedback 
	controllable Lipschitz control system, and let $S_\text{spec} 
	\negthinspace = \negthinspace (X_\text{spec},U_\text{spec}, 
	\ltsxition{S_\text{spec}\negthickspace\negthickspace}{~})$ be a transition 
	system encoding for a specification on $\Sigma$. Finally, let $\tau = 
	\tau(K_x,K_u,\mathcal{K},K_\text{cont},\delta)$ be determined by the 
	parameters specified.

	Now, suppose that there exists a $\delta$,$\tau$ positively invariant 
	Lipschitz-continuous controller $\Psi: \mathbb{R}^n \rightarrow U$ with 
	Lipschitz constant $K_\Psi \le K_\text{cont}$ such that:
	\begin{equation}
	\label{eq:main_prob_spec_assumtion}
		S_\tau(\Sigma_\Psi)
			\preceq_{\mathcal{AD}_\delta}
		S_\text{spec}.
	\end{equation}

	\noindent Then the problem is to find a ReLU architecture, 
	$\text{Arch}(\Theta)$, with the property that there exists values for 
	$\Theta$ such that:
	\begin{equation}\label{eq:main_prob_spec_conclusion}
		S_{\tau}(\Sigma_{\negthinspace\nn\negthinspace_\Theta})
			\preceq_{\mathcal{AD}_0}
		S_\text{spec}.
	\end{equation}
\end{problem}


The main assumption in \cref{prob:main_problem_generic} is that there exists a 
controller $\Psi$ which satisfies the specification, $S_\text{spec}$. We use 
this assumption largely to help ensure that the problem is well posed. For 
example, this assumption ensures that we aren't trying to assert the existence 
of NN controller for a system and specification that can't be achieved by 
\emph{any} continuous controller -- such examples are known to exist for 
nonlinear systems. In this way, the existence of a controller $\Psi$ subsumes 
any possible conditions of this kind that one might wish to impose: 
stabilizability, 
for example. Finally, note that the existence of such a $\Psi$ may require 
knowledge of $f$ to verify, but once its existence can be asserted the only 
explicit knowledge of $f$ we assume is $K_x$, $K_u$ and $\mathcal{K}$.

Moreover, there is a strong conceptual reason to consider abstract disturbance 
simulation in specification satisfaction for such a $\Psi$. Our approach to 
solve this problem will be to design a NN architecture that can approximate 
\textbf{\itshape any} such $\Psi$ sufficiently closely. However, $\nn_\Theta$ 
clearly belongs to a smaller class of functions than $\Psi$, so an arbitrary 
controller $\Psi$ cannot, in general, be represented \emph{exactly} by means of 
$\nn_\Theta$. This presents an obvious difficulty because instantaneous errors 
between $\Psi$ and $\nn_\Theta$ may accumulate by means of the system dynamics, 
i.e. via \eqref{eq:feedback_integral_eq}.

\section{ReLU Architectures for Nonlinear Systems} 
\label{sec:a_relu_architecture_for_nonlinear_systems}

Before we state the main theorem of the paper, we introduce the following 
notation in the form of a definition.

\begin{definition}[Extent of $X$]
	The \textbf{extent} of a compact set $X$ is defined as:
	\begin{equation}
		\text{ext}(X) \triangleq \max_{k=1,\dots,n} \left| 
		\max_{x \in X} \pi_k(x) - \min_{x\in X}\pi_k(x)
		\right|,
	\end{equation}
	where $\pi_k(x)$ is the projection of $x$ onto its $k^\text{th}$ component.
\end{definition}

The main result of the paper is the following theorem, which directly solves 
\cref{prob:main_problem_generic}.
\begin{theorem}[ReLU Architecture]\label{thm:main_theorem}
	Let $\delta > 0$ and $K_\text{cont} > 0$ be given, and let $\Sigma$ and 
	$S_\text{spec}$ be as in the statement of \cref{prob:main_problem_generic}. 
	Finally, choose a $\mu > 0$ such that:
	\begin{equation}\label{eq:main_inequality}
		K_u \cdot \mu \cdot \frac{\mu}{6 \cdot K_\text{cont} \cdot \mathcal{K}} \cdot
		e^{K_x \frac{\mu}{6 \cdot K_\text{cont} \cdot \mathcal{K}}} < \delta,
	\end{equation}
	and set:
	\begin{equation}
		\tau \leq \frac{\mu}{6\cdot K_\text{cont}\cdot 
			\mathcal{K} }
		\quad \text{ and } \quad
		\eta \leq \frac{\mu}{6 \cdot K_\text{cont}},
	\end{equation}
	(which depend only on $\mathcal{K}$, $K_x$, $K_u$, $K_\text{cont}$ and 
	$\delta$).

	If there exists a $\delta,\tau$ positively invariant Lipschitz continuous 
	controller $\Psi: \mathbb{R}^n \rightarrow U$ with Lipschitz constant 
	$K_\Psi \leq K_\text{cont}$ such that:
	\begin{equation}
	\label{eq:main_thm_spec_assumtion}
		S_\tau(\Sigma_\Psi)
			\preceq_{\mathcal{AD}_\delta}
		S_\text{spec}.
	\end{equation}
	Then a TLL NN architecture $\text{Arch}(\tllTheta_\text{N})$ of size: 
	\begin{equation}
		N \geq m \cdot
		\left(
			n! \cdot \sum_{k=1}^{n} \frac{2^{2k-1}}{(n-k)!}
		\right)
		\cdot
		\left(\frac{\text{ext}(X)}{\eta}\right)^n
	\end{equation}
	has the property that there exist values for $\tllTheta_\text{N}$ such that:
	\begin{equation}\label{eq:main_thm_spec_conclusion}
		S_{\tau}(\Sigma_{\negthinspace\nn\negthinspace_{\tllTheta_N}})
			\preceq_{\mathcal{AD}_0}
		S_\text{spec}.
	\end{equation}
\end{theorem}
\begin{remark}
	The left-hand side of \eqref{eq:main_inequality} looks like $\mu^2$ for 
	small $\mu$, so we can choose $\tau \approx \sqrt{\delta}$, thereby 
	obtaining $\lim_{\delta \rightarrow 0}\tfrac{\tau}{\delta} = +\infty$. Thus, 
	for $\delta$ small enough, the choice of $\tau$ in \cref{thm:main_theorem} 
	is compatible with a $\delta,\tau$ positively invariant controller (e.g. if 
	$\lVert f \rVert > a > 0$ on $\text{edge}_{\delta^\prime}(X)$ for some 
	$\delta^\prime > 0$).
\end{remark}
\noindent\textbf{Proof Sketch:}\\
The proof of \cref{thm:main_theorem} consists of establishing 
the following two implications:

\begin{enumerate}[{Step }1)]
	\item \emph{``Approximate controllers satisfy the specification'':} 
		There is an approximation accuracy, $\mu$, and sampling period, $\tau$, 
		with the following property: if the unknown controller $\Psi$ satisfies 
		the specification (under $\delta$ disturbance and sampling period 
		$\tau$), then any controller -- NN or otherwise -- which approximates 
		$\Psi$ to accuracy $\mu$ will also satisfy the specification (but under 
		no disturbance). 
		See \cref{lem:final_spec} of \cref{sec:thm1_part1}. 

	\item \emph{``Any controller can be approximated by a CPWA with the same 
		fixed number of linear regions'':} If unknown controller $\Psi$ has a 
		Lipschitz constant $K_\Psi \leq K_\text{cont}$, then $\Psi$ can be 
		approximated by a CPWA with a number of regions that depends only on 
		$K_\text{cont}$ and the desired approximation accuracy.
		See \cref{lem:cpwa_construction} of \cref{sec:thm1_part2}. 
\end{enumerate}

We will show these results for \emph{any} controller $\Psi$ that satisfies the 
assumptions of \cref{thm:main_theorem}. Thus, these results together show the 
following implication: if there \emph{exists} a controller $\Psi$ that 
satisfies the assumptions of \cref{thm:main_theorem}, then there is a CPWA 
controller that satisfies the specification. And moreover, this CPWA controller 
has at most a number of linear regions that depends only on the parameters of 
the problem \emph{and not the particular controller $\Psi$}.

The conclusion of the theorem then follows directly from 
\cref{thm:tll_architecture} \cite[Theorem 7]{FerlezAReNAssuredReLU2020}: 
together, they specify that any CPWA with the same number of linear regions (or 
fewer) can be implemented exactly by a common TLL NN architecture.
\begin{techreport}
Since this 
proof is so short given the lemmas described above, it appears in 
Appendix E of \cite{FerlezTwoLevelLatticeNeural2020}.
\end{techreport}
\begin{proofs}
Since this 
proof is so short given the lemmas described above, it appears in 
\cref{sub:proof_of_thm:main_theorem}.
\end{proofs}

\section{Proof of \cref{thm:main_theorem}, Step 1: Approximate Controllers Satisfy the Specification} 
\label{sec:thm1_part1}
The goal of this section is to choose constants $\mu > 0$ and $\tau > 0$ such 
that \emph{any} controller $\Upsilon$ with $\lVert \Upsilon - \Psi 
\rVert_\infty \leq \mu/3$ satisfies the specification:
\begin{equation}
	S_\tau(\Sigma_\Upsilon) \preceq_{\mathcal{AD}_0} S_\text{spec}.
\end{equation}
The approach will be as follows. First, we confine ourselves to a region in the 
state space on which the controller $\Psi$ doesn't vary much: the size of this 
region is determined entirely by the approximation accuracy, $\mu$, and the 
bound on the Lipschitz constant, $K_\text{cont}$. Then we confine the 
\emph{trajectories} of $\Sigma_\Psi$ to this region by bounding the duration of 
those trajectories, i.e. $\tau$. Finally, we feed these results into a 
Gr\"onwall-type bound to choose $\mu$. In particular, we choose $\mu$ small 
enough such that the error incurred by using $\Upsilon$ instead of $\Psi$ is 
within the disturbance robustness, $\delta$. From this we will conclude that 
$\Upsilon$ satisfies the specification whenever $\lVert \Upsilon - \Psi \rVert 
\leq \mu/3$. A road map of these steps is as follows.

\begin{itemize}
	\item Let $\mu$ be an approximation error. Then:
		\begin{enumerate}[{\itshape i)}]
			\item Choose $\eta = \eta(\mu)$ such that a Lipschitz function 
				with constant $K_\text{cont}$ doesn't vary by more than $\mu/3$ 
				between any two points that are $2\eta$ apart.

			\item Choose $\tau = \tau(\mu)$ such that $\lVert x - 
				\xi_{xv}(\tau) \rVert \leq \eta$ for any continuous open-loop 
				control $v$ (use $\lVert f \rVert \leq \mathcal{K}$).

			\item Use i) and ii) to conclude that $\lVert 
				\Upsilon(\zeta_{x\Upsilon}(t)) - \Psi(\zeta_{x\Psi}(t)) \rVert 
				\leq \lVert \Upsilon - \Psi \rVert_\infty + 2\mu/3$ for $t \in 
				[0,\tau]$

			\item Assume $ \lVert \Upsilon - \Psi \rVert_\infty \leq \mu/3$. 
				Choose $\mu = \mu(\delta)$ such that a Gr\"onwall-type bound 
				satisfies:
				\begin{multline}
					\lVert \zeta_{x\Upsilon}(\tau(\mu)) - \zeta_{x\Psi}(\tau(\mu))\rVert \leq \\
					K_u \cdot \mu \cdot \tau(\mu) \cdot e^{K_x\tau(\mu)} < \delta.
				\end{multline}
				Conclude that if $ \lVert \Upsilon - \Psi \rVert_\infty \leq 
				\mu/3$, then:
				\begin{equation}
					S_{\tau}( \Sigma_{\Upsilon} )
						\preceq_{\mathcal{AD}_0}
					\mathfrak{S}_{\tau}( \Sigma_{\Psi} )
						\preceq_{\mathcal{AD}_0}
					S_\text{spec}.
				\end{equation}
		\end{enumerate}
\end{itemize}
\begin{arxivreference}
Full proofs of the following can be found in 
\cite{FerlezTwoLevelLatticeNeural2020}.
\end{arxivreference}

\begin{proofs}
Now we proceed with the proof.
\end{proofs}
First, we formalize \emph{i)} - \emph{iii)} in the following propositions.
\begin{techreport}
The proofs of which are given in Appendix A - Appendix C of 
\cite{FerlezTwoLevelLatticeNeural2020}.
\end{techreport}
\begin{proofs}
The proofs of the propositions are given in 
\cref{sub:proof_of_prop:eta_proposition} - 
\cref{sub:proof_of_prop:closed_loop_control_bounds}.
\end{proofs}
\begin{proposition}\label{prop:eta_proposition}
	Let $\mu > 0$ be given, and let $\Psi$ be as above. Then there exists an 
	$\eta = \eta(\mu)$ such that:
	\begin{equation}
		\lVert x - x^\prime \rVert \leq 2 \eta
		\implies
		\lVert \Psi(x) - \Psi(x^\prime) \rVert \leq \mu/3.
	\end{equation}
\end{proposition}

\begin{proposition}\label{prop:tau_proposition}
	Let $\mu > 0$ be given, and let $\eta = \eta(\mu)$ be as in the previous 
	proposition. Finally, let $\Sigma$ be as specified in the statement of 
	\cref{thm:main_theorem}. Then there exists a $\tau = \tau(\mu)$ such that 
	for any Lipschitz feedback controller $\Upsilon$:
	\begin{equation}
		\lVert x - \zeta_{x\Upsilon}(t) \rVert \leq \eta = \eta(\mu) \quad \forall t \in [0, \tau].
	\end{equation}
\end{proposition}

\begin{proposition}
\label{prop:closed_loop_control_bounds}
	Let $\mu > 0$ be given. Let $\Sigma$ and $\Psi$ be as in the statement of 
	\cref{thm:main_theorem}; let $\eta = \eta(\mu)$ be as in 
	\cref{prop:eta_proposition}; let $\tau = \tau(\mu)$ be as in 
	\cref{prop:tau_proposition}; and let $\Upsilon : \mathbb{R}^n \rightarrow 
	U$ be a Lipschitz continuous function. Then:
	\begin{equation}
		\forall t \in [0,\negthinspace\tau] \thickspace\thickspace
		\lVert 
				\Upsilon(\zeta_{x\Upsilon}(t)) \negthinspace - \negthinspace
				\Psi(\zeta_{x\Psi}(t))
		\rVert
			\leq
		\lVert 
			\Upsilon \negthinspace- \negthinspace \Psi
		\rVert_\infty
		+
		\tfrac{2\mu}{3}
	\end{equation}
\end{proposition}

To prove Step \emph{iv)} we first need the following two results.
\begin{proposition}[Gr\"onwall Bound]
\label{prop:gronwall_lemma}
	Let $\Sigma$ and $\Psi$ be as in the statement of \cref{thm:main_theorem}, 
	and let $\Upsilon$ be as in the statement of 
	\cref{prop:closed_loop_control_bounds}. If:
	\begin{equation}
		\lVert 
				\Upsilon(\zeta_{x\Upsilon}(t)) - 
				\Psi(\zeta_{x\Psi}(t))
		\rVert \leq \kappa \quad \forall t \in [0,\tau]
	\end{equation}
	then:
	\begin{equation}
		\lVert \zeta_{x\Upsilon}(t) - \zeta_{x\Psi}(t) \rVert 
		\leq K_u \cdot \kappa \cdot t \cdot e^{K_x t}
		\quad
		\forall t \in [0,\tau].
	\end{equation}
\end{proposition}
\begin{techreport}
The proof of \cref{prop:gronwall_lemma} appears in 
Appendix D of \cite{FerlezTwoLevelLatticeNeural2020}.
\end{techreport}
\begin{proofs}
The proof of \cref{prop:gronwall_lemma} appears in 
\cref{sub:proof_of_prop:gronwall_lemma}.
\end{proofs}

\begin{lemma}
\label{lem:delta_state_bound}
	Let $\Sigma$, $\Psi$ and $\Upsilon$ be as before. Also, suppose that $\mu > 
	0$ is such that:
	\begin{equation}
		K_u \cdot \mu \cdot \frac{\mu}{6 \cdot K_\text{cont} \cdot \mathcal{K}} \cdot
		e^{K_x \frac{\mu}{6 \cdot K_\text{cont} \cdot \mathcal{K}}} < \delta.
	\end{equation}
	If $\lVert \Upsilon - \Psi \rVert_\infty \leq \mu/3$, then:
	\begin{equation}
		\lVert \zeta_{x\Upsilon}(\tau(\mu)) - \zeta_{x\Psi}(\tau(\mu))\rVert \leq \delta.
	\end{equation}
\end{lemma}
\begin{proofs}
\begin{proof}
	This is a direct consequence of applying 
	\cref{prop:closed_loop_control_bounds} to \cref{prop:gronwall_lemma}.
\end{proof}
\end{proofs}

\noindent The final result in this section is the following Lemma.
\begin{lemma}
\label{lem:final_spec}
	Let $\Sigma$, $\Psi$ and $\Upsilon$ be as before, and suppose that $\mu > 
	0$ is such that:
	\begin{equation}
		K_u \cdot \mu \cdot \frac{\mu}{6 \cdot K_\text{cont} \cdot \mathcal{K}} \cdot
		e^{K_x \frac{\mu}{6 \cdot K_\text{cont} \cdot \mathcal{K}}} < \delta.
	\end{equation}
	If $\lVert \Upsilon - \Psi \rVert_\infty \leq \mu/3$, then for $\tau \leq 
	\tfrac{\mu}{6\cdot K_\text{cont} \cdot \mathcal{K}}$ we have:
	\begin{equation}
		S_{\tau}( \Sigma_{\Upsilon} )
			\preceq_{\mathcal{AD}_0}
		\mathfrak{S}_{\tau}( \Sigma_{\Psi} ).
	\end{equation}
	And hence:
	\begin{equation}
		S_{\tau}( \Sigma_{\Upsilon} ) \preceq_{\mathcal{AD}_0} S_\text{spec}.
	\end{equation}
\end{lemma}
\begin{proofs}
\begin{proof}
	 By definition, $S_{\tau}( \Sigma_{\Upsilon} )$ and $\mathfrak{S}_{\tau}( 
	\Sigma_{\Psi} )$ have the same state spaces, $X$. Thus we propose the 
	following as an abstract disturbance simulation under $0$ disturbance (i.e. 
	a conventional simulation for metric transition systems):
	\begin{equation}
		R = \{(x,x) | x \in X\}.
	\end{equation}

	Clearly, $R$ satisfies the property that for all $(x,y) \in R$, $d(x,y) 
	\leq 0$, and for every $x\in X$, there exists an $y \in X$ such that $(x,y) 
	\in R$. Thus, it only remains to show the third property of 
	\cref{def:delta_perturbation_simulation} under $0$ disturbance.

	To wit, let $(x,x) \in R$. Then suppose that $x^\prime \triangleq 
	\zeta_{x\Upsilon}(\tau) \in X$, so that $x 
	\xition{\Upsilon}{\Upsilon\circ\zeta_{x\Upsilon}} x^\prime$ in $S_{\tau}( 
	\Sigma_{\Upsilon} )$; we will show subsequently that any such $x^\prime$ 
	must be in $X$. In this situation, it suffices to show that $x 
	\xition{{\Psi}}{\Psi\circ\zeta_{x\Psi}} x^\prime$ in $\mathfrak{S}_{\tau}( 
	\Sigma_{\Psi} )$. 
	By the $\delta, \tau$ positive invariance of $\Psi$, it is the case that 
	$x^{\prime\prime} \triangleq \zeta_{x\Psi}(\tau) \in X \backslash 
	\text{edge}_\delta(X)$, so $x \xition{{\Psi}}{\Psi\circ\zeta_{x\Psi}} 
	x^{\prime\prime}$ in $S_\tau(\Sigma_\Psi)$. But by 
	\cref{lem:delta_state_bound}, $\lVert x^\prime - x^{\prime\prime} \rVert 
	\leq \delta$, so $x \xition{{\Psi}}{\Psi\circ\zeta_{x\Psi}} x^\prime$ in 
	$\mathfrak{S}_{\tau}( \Sigma_{\Psi} )$ by definition.

	It remains to show that any such $x^\prime \triangleq 
	\zeta_{x\Upsilon}(\tau) \in X$. This follows by contradiction from 
	\cref{lem:delta_state_bound} and the $\delta, \tau$ positive invariance of 
	$\Psi$. That is suppose $x^\prime \notin X$. By 
	\cref{lem:delta_state_bound}, $x^{\prime\prime} \triangleq 
	\zeta_{x\Psi}(\tau)$ satisfies $\lVert x^\prime - x^{\prime\prime} \rVert 
	\leq \delta$, which implies that $x^{\prime\prime} \in 
	\text{edge}_\delta(X) \cup \mathbb{R}^n \backslash X$. However, this clearly 
	contradicts the $\delta,\tau$ positive invariance of $\Psi$.
\end{proof}
\end{proofs}

\section{Proof of \cref{thm:main_theorem}, Step 2: CPWA Approximation of a Controller} 
\label{sec:thm1_part2}
The results in \cref{sec:thm1_part1} showed that any controller, $\Upsilon$, 
whether it is CPWA or not, will satisfy the specification if it is close to 
$\Psi$ in the sense that $\lVert \Upsilon - \Psi \rVert_\infty \leq \mu/3$ 
(where $\mu$ is as specified therein). Thus, the main objective of this section 
will be to show that an arbitrary $\Psi$ can be approximated to this accuracy 
by a CPWA controller, $\Upsilon\subcpwa$, subject to the following caveat. It 
is well known that CPWA functions are good function approximators in general, 
but we have to keep in mind our eventual use of \cref{thm:tll_architecture}: we 
need to approximate \emph{any such} $\Psi$ by a CPWA with the \textbf{same, 
bounded number of linear regions}. Hence, our objective in this section is to 
find not just a controller $\Upsilon\subcpwa$ that approximates $\Psi$ to the 
specified accuracy, but one that achieves this using not more than some common, 
fixed number of linear regions that depends only on the problem parameters (and 
not the function $\Psi$ itself, which is assumed unknown except for a bound on 
its Lipschitz constant).

With this in mind, our strategy will be to partition the set $X$ into a grid of 
$\sup$-norm balls such that no relevant $\Psi$ can vary by much between them: 
indeed, we will use balls of size $\eta$, as specified in 
\cref{sec:thm1_part1}. Thus, we propose the following starting point: inscribe 
a slightly smaller ball within each $\eta$ ball of the partition, and choose 
the value of $\Upsilon\subcpwa$ on each such ball to be a constant value equal 
to $\Psi(x)$ for some $x$ therein. Because we have chosen the size of the 
partition to be small, such an $\Upsilon\subcpwa$ will still be a good 
approximation of $\Psi$ for these points in its domain. Using this approach, 
then, we only have to concern ourselves with how ``extend'' a function so 
defined to the entire set $X$ as a CPWA. Moreover, note that this procedure is 
actually independent of the particular $\Psi$ chosen, despite appearances: we 
are basing our construction on a grid size $\eta$ that depends only on the 
problem parameters (via $\mu$), so the construction will work no matter the 
chosen value of $\Psi(x)$ within each grid square.

The first step in this procedure will be to show how to extend such a function 
over the largest-dimensional ``gaps'' between the smaller inscribed balls 
\begin{proofs}
; the blue region depicted in \cref{fig:cpwa_interp_notation} is an example of 
this large-dimensional gap for $X \subset \mathbb{R}^2$ (the notation in the 
figure will be explained later)\hspace{0.3em}
\end{proofs}
\hspace{-0.3em}. This result must control the error of the extension so as to 
preserve our desired approximation bound, as well provide a count of the number 
of linear regions necessary to do so; this is \cref{lem:cpwa_interpolate}. The 
preceding result can then be extended to all of the other gaps between 
inscribed balls to yield a CPWA function with domain $X$, approximation error 
$\mu/3$, and a known number of regions; this is \cref{lem:cpwa_construction}.
\begin{arxivreference}
Full proofs appear in \cite{FerlezTwoLevelLatticeNeural2020}.
\end{arxivreference}

\begin{proofs}
\begin{definition}[Face]
	Let $C = [0,1]^n$ be a unit hypercube of dimension $n$. A set $F \subseteq 
	C$ is a $k$-dimensional \textbf{face} of $C$ if there exists a set $J 
	\subseteq \{1, \negthinspace \dots, \negthinspace n\}$ such that $|J| = n 
	\negthinspace - \negthinspace k$ and
	\begin{equation}
		\forall x \in F ~.~ \bigwedge_{j \in J} \pi_j(x) \in \{0,1\}.
	\end{equation}
	Let $\mathscr{F}_k(C)$ denote the set of $k$-dimensional faces of $C$, and 
	let $\mathscr{F}(C)$ denote the set of all faces of $C$ (of any dimension).
\end{definition}
\begin{remark}
	A $k$-dimensional face of the hypercube $C = [0,1]^n$ is isomorphic to the 
	hypercube $[0,1]^k$.
\end{remark}
\begin{definition}[Corner]
	Let $C = [0,1]^n$. A \textbf{corner} of $C$ is a $0$-dimensional face of 
	$C$.
\end{definition}
\end{proofs}
\begin{arxivreference}
\begin{definition}[Face/Corner]
	Let $C = [0,1]^n$ be a unit hypercube of dimension $n$. A set $F \subseteq 
	C$ is a $k$-dimensional \textbf{face} of $C$ if there exists a set $J 
	\subseteq \{1, \negthinspace \dots, \negthinspace n\}$ such that $|J| = n 
	\negthinspace - \negthinspace k$ and
	\begin{equation}
		\forall x \in F ~.~ \bigwedge_{j \in J} \pi_j(x) \in \{0,1\}.
	\end{equation}
	Let $\mathscr{F}_k(C)$ denote the set of $k$-dimensional faces of $C$, and 
	let $\mathscr{F}(C)$ denote the set of all faces of $C$ (of any dimension). 
	A \textbf{corner} of $C$ is a $0$-dimensional face of $C$.
\end{definition}
\end{arxivreference}

\begin{lemma}
\label{lem:cpwa_interpolate}
	Let $C = [0,1]^n$, and suppose that:
	\begin{equation}
		\Gamma_c : \mathscr{F}_0(C) \rightarrow \mathbb{R}
	\end{equation}
	is a function defined on the corners of $C$. Then there is a CPWA function 
	$\Gamma: C \rightarrow \mathbb{R}$ such that:
	\begin{itemize}
		\item $\forall x \in \mathscr{F}_0(C). \Gamma(x) = \Gamma_c(x)$, 
			i.e. $\Gamma$ extends $\Gamma_c$ to $C$;

		\item $\Gamma$ has at most $2^{n-1} \cdot n!$ linear regions; and

		\item for all $x \in C$,
			\begin{equation}
				 \min_{x \in \mathscr{F}_0(C)} \Gamma_c(x) 
				\leq \Gamma(x) \leq \max_{x \in \mathscr{F}_0(C)} \Gamma_c(x).
			\end{equation}
	\end{itemize}
\end{lemma}
\begin{proofs}
\begin{proof}
	First, we assume without loss of generality that the given function 
	$\Gamma_c$ takes distinct values on each element of its domain.

	This is a proof by induction on dimension. In particular, we will use the 
	following induction hypothesis:
	\begin{itemize}
		\item There is a function $\Gamma_k : \cup_{i=1}^k \mathscr{F}_i(C) 
			\rightarrow \mathbb{R}$ such that for all $\tilde{F} \in 
			\mathscr{F}_k(C)$, $\Gamma_k |_{\tilde{F}}$ has the following 
			properties:
			\begin{itemize}
				\item it is CPWA

				\item it has at most $2^{k-1} \cdot k!$ linear regions; and

				\item for all $x\in \tilde{F}$:
					\begin{equation}
						\min_{x\in \mathscr{F}_0(\tilde{F})} \Gamma_c(x)
						\leq
						\Gamma_k(x)
						\leq
						\max_{x\in \mathscr{F}_0(\tilde{F})} \Gamma_c(x).
					\end{equation}
			\end{itemize}
	\end{itemize}
	We start by showing that if the induction hypothesis above holds for $k$, 
	then it also holds for $k+1$.

	To show the induction step, first note that for any face $F \in 
	\mathscr{F}_{k+1}(C)$, all of \emph{its} faces are already in the domain of 
	$\Gamma_k$. That is $\cup_{i=1}^k \mathscr{F}_i(F) \subseteq 
	\text{dom}(\Gamma_k)$. Thus, we can define $\Gamma_{k+1}$ by extending 
	$\Gamma_k$ to $\text{int}(F)$ for each $F \in \mathscr{F}_{k+1}(C)$. Since 
	these interiors are mutually disjoint, we can do this by explicit 
	construction on each individually, in such a way that the desired 
	properties hold.

	In particular, let $F \in \mathscr{F}_{k+1}(C)$, and let $\nu$ be the 
	midpoint of $F$, i.e. the $k$-cube isomorphism of $\nu$ is $[\tfrac{1}{2}, 
	\dots, \tfrac{1}{2}]$. $\nu$ is clearly in the interior of $F$, so  define:
	\begin{equation}
		\Gamma_{k+1}(\nu) = \frac{1}{|\mathscr{F}_0(F)|} \sum_{x \in \mathscr{F}_0(F)} \Gamma_k(x)
	\end{equation}
	and note that the corners of $F$ are also corners of $C$ Thus, 
	$\Gamma_{k+1}(\nu)$ is the average of all of the corners of the $k+1$-face 
	that contains it. Now, extend $\Gamma_{k+1}$ to the rest of $\text{int}(F)$ 
	as follows: let $b \in \cup_{i=1}^k \mathscr{F}_i(F)$ and define:
	\begin{multline}
		\Gamma_{k+1}(\lambda\cdot \nu + (1-\lambda) \cdot b)
		= \\
		\lambda \cdot \Gamma_{k+1}(\nu) + (1-\lambda) \cdot \Gamma_k(b)
		\quad
		\forall \lambda \in [0,1].
	\end{multline}
	This definition clearly covers $\text{int}(F)$, and it also satisfies the 
	requirement that:
	\begin{equation}
		\forall x \in F ~
		\min_{x \in \mathscr{F}_0(F)} \Gamma_c(x)
		\leq
		\Gamma_{k+1}(x)
		\leq
		\max_{x \in \mathscr{F}_0(F)} \Gamma_c(x)
	\end{equation}
	because the induction hypothesis ensures that each $b$ is on a face of $F$, 
	and the corners of a face of $F$ are a subset of the corners of $F$. Thus, 
	it remains to show the bound on the number of linear regions. But from the 
	construction, $\Gamma_{k+1}|_F$ has one linear region for linear region of 
	$\Gamma_k$ on a $k$-face of $F$. Since the $k+1$-face $F$ has $2 \cdot 
	(k+1)$ $k$-faces, we conclude by the induction hypothesis that 
	$\Gamma_{k+1}|_F$ has at most:
	\begin{equation}
		2 \cdot (k+1) \cdot 2^{k-1} \cdot k! = 2^k \cdot (k+1)!
	\end{equation}
	linear regions. This completes the proof of the induction step.

	It remains only to show a base case. For this, we select $k=1$, i.e. the 
	line-segment faces of $C$. Each $1$-face of C has only two corners and no 
	other faces other than itself. Thus, for each $F \in \mathscr{F}_0(C)$ we 
	can simply define $\Gamma_1|_F$ to linearly interpolate between those two 
	corners. $\Gamma_1|_F$ is thus CPWA, and it satisfies the required bounds 
	on its values. Moreover, $\Gamma_1|_F$ has exactly $2^{1-1} \cdot 1! = 1$ 
	linear region. Thus, the function $\Gamma_1$ so defined satisfies the 
	induction hypothesis stated above.
\end{proof}
\end{proofs}

\begin{proofs}
\begin{definition}[$\eta$-partition]
	Let $\eta > 0$ be given. Then an $\eta$-partition of $X$ is a regular, 
	non-overlapping grid of $\eta/2$ balls in the $\sup$ norm that partitions 
	$X$. Let $X_\text{cent}$ denote the set of centers of these balls, and let 
	$X_\text{part} = \{B(x_c;\eta/2) | x_c \in X_\text{cent}\}$ denote the 
	partition.
\end{definition}

\begin{definition}[Neighboring Grid Center/Square]
	Let $X_\text{part}$ be an $\eta$-partition of $X$, and let $B(x_c;\eta/2) 
	\in X_\text{part}$. Then a \textbf{neighboring grid center (resp. square)} 
	to $x_c$ is an $x_c^\prime \in X_c$ (respectively $B(x_c^\prime;\eta/2) \in 
	X_\text{part}$) such that $B(x_c^\prime;\eta/2)$ shares a face (of any 
	dimension) with $B(x_c;\eta/2)$. The set of neighbors of a center, $x_c$, 
	will be denoted by $\mathcal{N}(x_c)$.
\end{definition}
\end{proofs}

\begin{lemma}
\label{lem:cpwa_construction}
	Let $\eta = \eta(\mu)$ be chosen as in \cref{prop:eta_proposition}, and let 
	$\Psi$ be as before. Then there is a CPWA function $\Upsilon\subcpwa : 
	\mathbb{R}^n \rightarrow U$ such that:

	\begin{itemize}
		\item $\lVert \Upsilon\subcpwa - \Psi \rVert_\infty \leq 
			\tfrac{\mu}{3}$; and

		\item $\Upsilon\subcpwa$ has linear regions numbering at most
			\begin{equation}\label{eq:region_count}
				m \cdot
				\left( 
					n! \cdot \sum_{k=1}^{n} \frac{2^{2k-1}}{(n-k)!}
				\right)
				\cdot 
				\left(
					\frac{\text{ext}(X)}{\eta}
				\right)^n.
			\end{equation}
	\end{itemize}
\end{lemma}

\begin{proofs}
\begin{proof}
	Our proof will assume that $U \subseteq \mathbb{R}$, since the extension to 
	$m > 1$ is straightforward from the $m = 1$ case. The basic proof will be 
	to create an $\eta$-partition of $X$, and define $\Upsilon\subcpwa$ to be 
	constant on $\rho \cdot \eta/2 < \eta/2$ radius balls centered at each of 
	the grid centers in the partition; we will then use 
	\cref{lem:cpwa_interpolate} to ``extend'' this function to the rest of $X$ 
	as a CPWA function. In particular, for each $x_c \in X_C$, we start by 
	defining:
	\begin{equation}
		\Upsilon\subcpwa(x) = \Psi(x_c) \quad \forall x \in B(x_c; \rho\cdot \eta/2).
	\end{equation}
	Then we will extend this function to the rest of $X$, and prove the claims 
	for that extension.

	To simplify the proof, we will henceforth focus on a particular $x_c$, and 
	show how to extend $\Upsilon\subcpwa$ from $B(x_c; \rho\cdot \eta/2)$ to the 
	``gaps'' between it and each of the neighboring balls, $B(x_c^\prime; 
	\rho\cdot \eta/2)$ for $x_c^\prime \in \mathcal{N}(x_c)$. To further 
	simplify the proof, we define here two additional pieces of notation. 
	First, for each $x_c \in X_c$ and each $k \in \{1,\dots,n\}$ define a 
	function $\omega_k^{(x_c)}$ as follows:
	\begin{align}
		\omega_k^{(x_c)} &: \{-1, 0, +1\} \rightarrow 2^\mathbb{R} \notag\\
		\omega_k^{(x_c)} &: \hphantom{+}0            \mapsto     
				[\pi_k(x_c) - \rho\tfrac{\eta}{2} ,~ \pi_k(x_c) + \rho\tfrac{\eta}{2}] \notag \\
		\omega_k^{(x_c)} &: +1            \mapsto     
				[\pi_k(x_c) + \rho\tfrac{\eta}{2}, \pi_k(x_c) + \tfrac{\eta}{2} + (1-\rho)\tfrac{\eta}{2}] \notag \\
		\omega_k^{(x_c)} &: -1            \mapsto     
				[\pi_k(x_c) - \tfrac{\eta}{2} - (1-\rho)\tfrac{\eta}{2}, \pi_k(x_c) - \rho\tfrac{\eta}{2}]. \notag
	\end{align}
	Then, define the function: 
	\begin{align}
		\mathcal{R}^{(x_c)} &: \{-1,0,1\}^n \rightarrow {2^\mathbb{R}}^n \notag \\
		\mathcal{R}^{(x_c)} &: \iota \mapsto
			\omega_1^{(x_c)}(\pi_1(\iota)) \times \omega_2^{(x_c)}(\pi_2(\iota)) \times \dots \times \omega_n^{(x_c)}(\pi_n(\iota)), \notag
	\end{align}
	and let $\mathbf{0} \triangleq (0,\dots,0) \in \{-1,0,1\}^n$. Also, define 
	$\text{dim}(\iota)$ as the number of non-zero elements in $\iota$.

	Now let $x_c \in X_c$ be fixed. Using the above notation, the ball $B(x_c; 
	\rho\cdot \eta/2)$ is given by: 
	\begin{equation}
		B(x_c; \rho\cdot \eta/2) = \mathcal{R}^{(x_c)}(\mathbf{0}),
	\end{equation}
	Similarly each of the ``gaps'' between $\mathcal{R}^{(x_c)}(\textbf{0})$ 
	and its neighbors, $\mathcal{R}^{(x_c^\prime)}(\mathbf{0})$ for $x_c^\prime 
	\in\mathcal{N}(x_c)$, are the hypercubes: 
	\begin{equation}
		\mathcal{R}^{(x_c)}(\iota)
		\text{ for } \iota \in \{-1,0,1\}^n\backslash \{\mathbf{0}\},
	\end{equation}
	and hence:
	\begin{multline}
		\bigcup_{\substack{x_c \in X_c, \\ \iota \in \{-1,0,1\}^n\backslash \mathbf{0}}} 
	\mathcal{R}^{(x_c)}(\iota) = 
	X \backslash \bigcup_{x_c^\prime \in X_c} B(x_c^\prime; \rho \cdot \eta/2 ).
	\end{multline}
	This notation is illustrated in two dimensions in 
	\cref{fig:cpwa_interp_notation}.

	\begin{figure}
		\centering
		\includegraphics[width=0.25\textwidth]{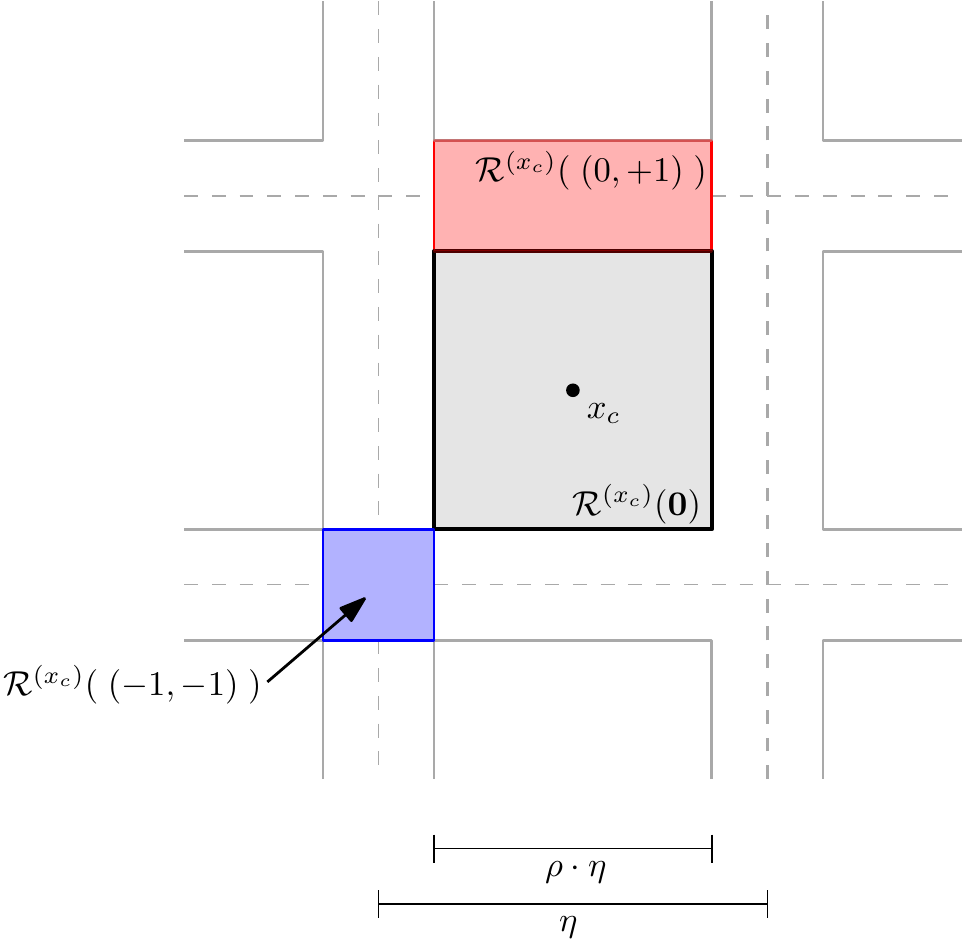} \vspace{-4mm}
		\caption{Illustration of $\mathcal{R}^{(x_c)}$ notation for $X \subset \mathbb{R}^2$. For $x_c$ as labeled, the regions $\mathcal{R}^{(x_c)}((-1,-1))$, $\mathcal{R}^{(x_c)}((0,+1))$ and $\mathcal{R}^{(x_c)}(\mathbf{0})$ are shown in blue, red and light gray, respectively.} \vspace{-7mm}
		\label{fig:cpwa_interp_notation}
	\end{figure}

	The first step is to show that $\Upsilon\subcpwa$ can be extended from the 
	constant-valued region, $\mathcal{R}^{(x_c)}(\mathbf{0})$, to each of its 
	neighbors, $\mathcal{R}^{(x_c)}(\iota)$, in a consistent way as a CPWA. To 
	do this, first note that $\mathcal{R}^{(x_c)}(\mathbf{0})$ has $2^n$ 
	neighboring regions with indices $\iota^\prime \in \{-1,+1\}^n$, and each 
	of these regions intersects a different 
	$\mathcal{R}^{(x_c^\prime)}(\mathbf{0})$ for $x_c^\prime \in 
	\mathcal{N}(x_c)$ at each corner, but is otherwise disjoint from them. 
	Thus, \cref{lem:cpwa_interpolate} can be used to define a CPWA on each such 
	$\mathcal{R}^{(x_c)}(\iota^\prime)$ in a way that is consistent with the 
	definition of $\Upsilon\subcpwa$ on the 
	$\mathcal{R}^{(x_c^\prime)}(\mathbf{0})$. These definitions are also 
	consistent with each other, since these regions are disjoint. Moreover, 
	this procedure yields the same extension when started from $x_c^\prime \in 
	\mathcal{N}(x_c)$ instead of $x_c$ (by the symmetric way that 
	\cref{lem:cpwa_interpolate} is proved). Thus, it remains only to define 
	$\Upsilon\subcpwa$ on regions with indices of the form $\iota^{\prime\prime} 
	\in \{-1,0,+1\}^n \backslash \{-1,+1\}^n \cup \{\mathbf{0}\}$. However, 
	each such $\mathcal{R}^{(x_c)}(\iota^{\prime\prime})$ intersects 
	$2^{n-\text{dim}(\iota^{\prime\prime})}$ regions with indices of the form 
	$\iota^\prime \in \{-1,+1\}^n$, and each of those intersections is a 
	$\text{dim}(\iota^{\prime\prime})$ face of the corresponding 
	$\mathcal{R}^{(x_c)}(\iota^\prime)$. But on each such 
	$\text{dim}(\iota^{\prime\prime})$ face, $\Upsilon\subcpwa$ is defined and 
	agrees with $\Gamma_{\text{dim}(\iota^{\prime\prime})}$ from the 
	construction in \cref{lem:cpwa_interpolate}. Finally, since 
	$\Gamma_{\text{dim}(\iota^{\prime\prime})}$ (and hence $\Upsilon\subcpwa$) is 
	identical up to isomorphism on each of these 
	$\text{dim}(\iota^{\prime\prime})$ faces, $\Upsilon\subcpwa$ can be extended on 
	to $\mathcal{R}^{(x_c)}(\iota^{\prime\prime})$ by isomorphism between the 
	$\text{dim}(\iota^{\prime\prime})$ nonzero indices, and $\Upsilon\subcpwa$ as 
	defined on one of the $\text{dim}(\iota^{\prime\prime})$ faces of 
	$\mathcal{R}^{(x_c)}(\iota^{\prime})$. Finally, the symmetry of this 
	procedure and \cref{lem:cpwa_interpolate} ensures that this assignment will 
	be consistent when starting from some $x_c^\prime \in \mathcal{N}(x_c)$ 
	instead of $x_c$.

	Next, we show that for this $\Upsilon\subcpwa$, $\lVert \Upsilon\subcpwa - 
	\Psi \rVert \leq \mu/3$. This largely follows from the interpolation 
	property proven in \cref{lem:cpwa_interpolate}. In particular, on some 
	$\mathcal{R}^{(x_c)}(\iota)$, $\Upsilon\subcpwa$ takes exactly the same 
	values as some $\Gamma_{\text{dim}(\iota)}$ constructed according to 
	\cref{lem:cpwa_interpolate}, where the interpolation happens between 
	$\text{dim}(\iota)$ points in $V \triangleq \{ \Psi(x_c^\prime) | 
	x_c^\prime \in \mathcal{N}(x_c) \cup \{x_c\} \}$. Thus,
	\begin{equation}
		\forall x \in \mathcal{R}^{(x_c)}(\iota)\quad
		\min_{y\in V} \Psi(y) \leq
		\Upsilon\subcpwa(x) \leq
		\max_{y \in V} \Psi(y).
	\end{equation}
	Let $x \in \mathcal{R}^{(x_c)}(\iota)$ be fixed temporarily, and suppose 
	that $\Upsilon\subcpwa(x) - \Psi(x) \geq 0$ and $\max_{y\in V}\Psi(y) - 
	\Psi(x) \geq 0$. Then:
	\begin{multline}
		|\Upsilon\subcpwa(x) - \Psi(x)| = \Upsilon\subcpwa(x) - \Psi(x) \\
		\leq
		\max_{y \in V} \Psi(y) - \Psi(x) 
			= |\max_{y \in V} \Psi(y) - \Psi(x)| \leq \frac{\mu}{3}
	\end{multline}
	where the last inequality follows from our choice of $\eta$ from 
	\cref{prop:eta_proposition}, since $|y - x| \leq 2 \eta$ for all $y \in V$. 
	The other cases can be considered as necessary, and they lead to the same 
	conclusion. Hence, we conclude $\lVert \Upsilon\subcpwa - 
	\Psi\rVert_\infty \leq \mu/3$, since our choice of center $x_c$ and $\iota$ 
	was arbitrary.

	Now we just need to (over)-count the number of linear regions needed in the 
	extension $\Upsilon\subcpwa$. This too will follow from the construction in 
	\cref{lem:cpwa_interpolate}. Note that on each 
	$\mathcal{R}^{(x_c)}(\iota)$, $\Upsilon\subcpwa$ has the same number of 
	linear regions as some $\Gamma_{\text{dim}(\iota)}$ that was constructed by 
	\cref{lem:cpwa_interpolate}, which by the same lemma has 
	$2^{\text{dim}(\iota) - 1} \cdot \text{dim}(\iota)!$ regions. Thus, we 
	count at most:
	\begin{equation}\label{eq:proof_region_count}
		\sum_{k=1}^n {n \choose k} \cdot 2^k \cdot 2^{k-1} \cdot k! = n! \cdot \sum_{k=1}^{n} \frac{2^{2k-1}}{(n-k)!}
	\end{equation}
	linear regions. Finally, since we need this many regions for the 
	neighboring regions of a single grid square, we obtain an upper bound for 
	the total number of regions by multiplying \eqref{eq:proof_region_count} by 
	the number of grid squares in the partition, 
	$(\tfrac{\text{ext}(X)}{\eta})^n$ (then by the $m$, in the 
	multi-dimensional output case).
\end{proof}
\end{proofs}




 %

\section{Numerical Results} 
\label{sec:numerical_results}
%

\begin{figure*}
    \vspace{1.4mm}
    \begin{tabular}{ccc}
         \includegraphics[width=0.29\textwidth, trim={0 0 0 0}, clip]{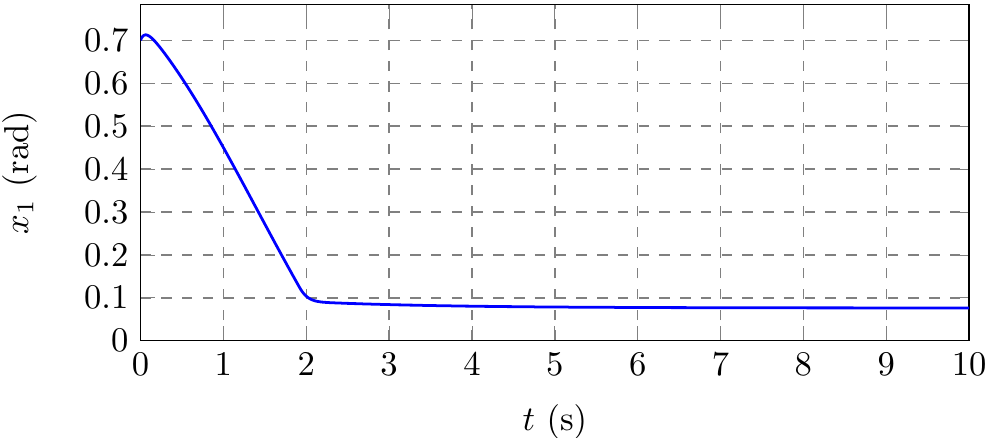} &\includegraphics[width=0.29\textwidth, trim={0 0 0 0}, clip]{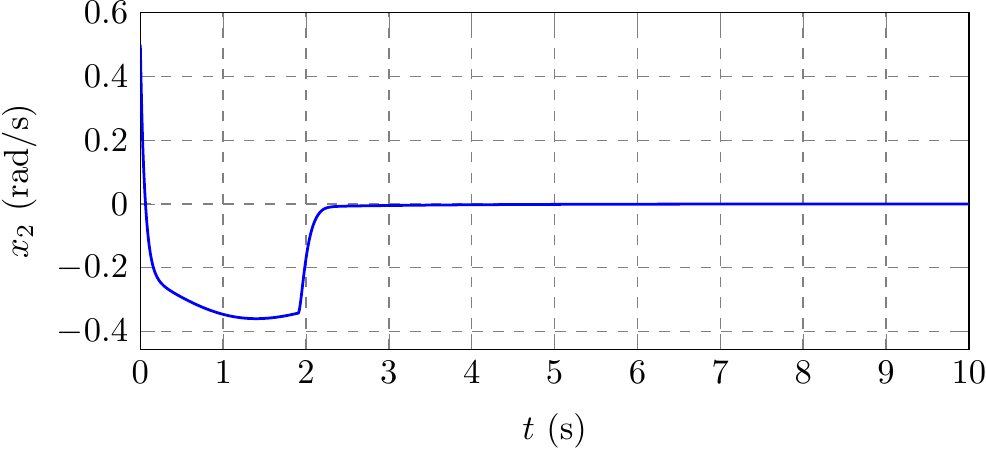} &\includegraphics[width=0.29\textwidth, trim={0 0 0 0}, clip]{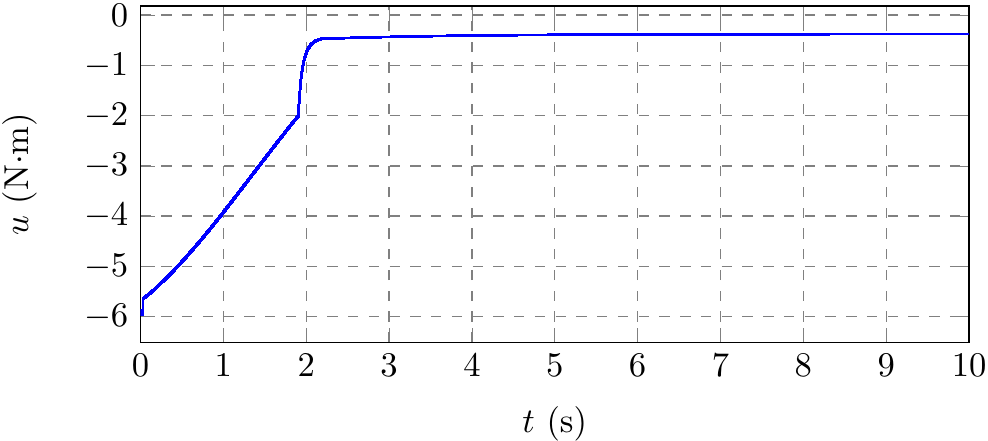}
    \end{tabular} \vspace{-3.5mm}
    \caption{States and inputs of the inverted pendulum with initial condition $[0.7, 0.5]^T$.} 
    \label{fig:x0p7_v0p5} \vspace{-3.5mm}
\end{figure*}

\begin{figure*}
    \begin{tabular}{ccc}
         \includegraphics[width=0.29\textwidth, trim={0 0 0 0}, clip]{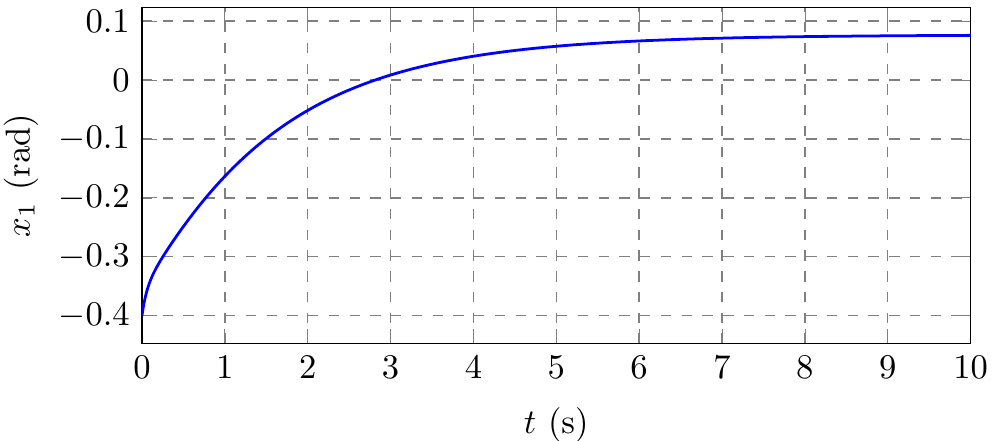} &\includegraphics[width=0.29\textwidth, trim={0 0 0 0}, clip]{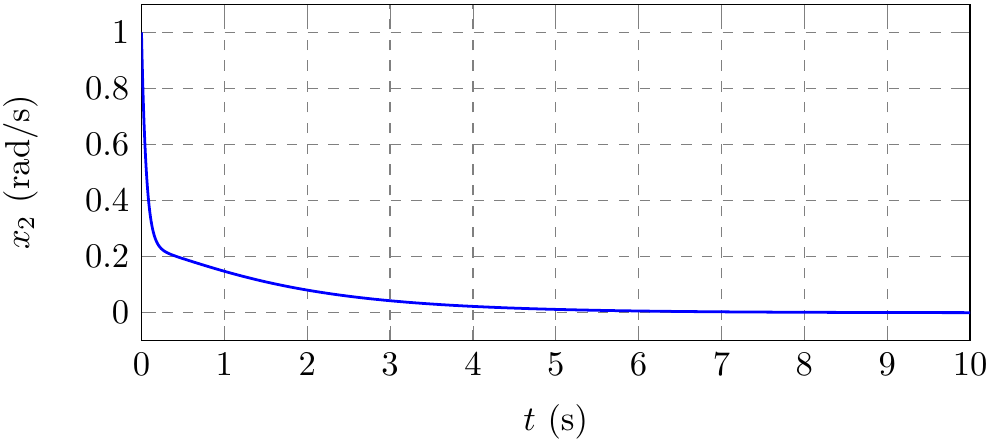} &\includegraphics[width=0.29\textwidth, trim={0 0 0 0}, clip]{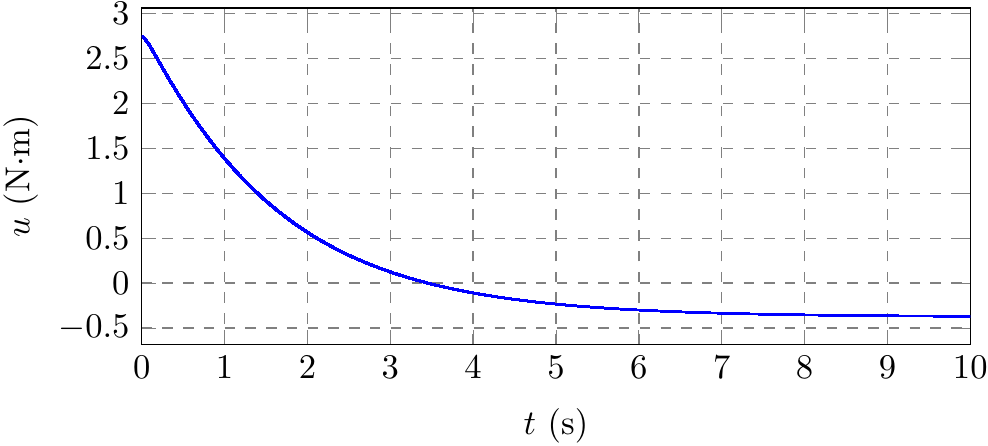}
    \end{tabular} \vspace{-3.5mm}
    \caption{States and inputs of the inverted pendulum with initial condition $[-0.4, 1.0]^T$} 
    \label{fig:xm0p4_v1}  \vspace{-7.5mm}
\end{figure*}

We illustrate the results in this paper on an inverted pendulum described by the following model:
\begin{equation*}
    f(x_1,x_2,u) = \left[\begin{matrix}
        x_2 &
        \frac{g}{l} \text{sin} (x_1) - \frac{h}{ml^2}x_2 + \frac{1}{ml} \text{cos}(x_1) u
    \end{matrix}\right]^\text{T},
\end{equation*}
where $x_1$ is the angular position, $x_2$ is the angular velocity, and control 
input $u$ is the torque applied on the point mass. The parameters are the rod 
mass, $m$; the rod length, $l$; the (dimensionless) coefficient of rotational 
friction, $h$; and the acceleration due to gravity, $g$. For the purposes of 
our experiments, we considered a subset of the state/control space specified 
by: $x_1 \in [-1, 1]$, $x_2\in [-1, 1]$ and $u \in [-6, 6]$. Furthermore, we 
considered model parameters: $m=0.5$ kg; $l=0.5$ m; $h=2$; and $g=9.8$ N/kg. 
Then for different choices of the design parameters $\mu$, we obtained the 
following table of sizes, $N$, for the corresponding TLL-NN architecture; also 
shown are the corresponding $\tau$, $\eta$ and the $\delta$ that are required 
for the specification satisfaction. \vspace{-0.75mm}
\begin{center}
\begin{tabular}{|c|c|c|c|c|}
\hline
$\mu$ & $\delta$ & $\tau$ & $\eta$ & $N$ \\ \hline
 0.35 & 0.8694 & 0.0098 & 0.583 & 235 \\ \hline
 0.3 & 0.5287 & 0.0083 & 0.5 & 320 \\ \hline
0.25 & 0.3039 & 0.0069 & 0.417 & 460 \\ \hline
0.2 & 0.1610 & 0.0056 & 0.334 & 720  \\ \hline
0.15 & 0.0749 & 0.0042 & 0.25 & 1280 \\ \hline
0.1 & 0.0275 & 0.0028 & 0.167 & 2880 \\ \hline
\end{tabular}
\end{center}\vspace{-0.75mm}

In the sequel, we will show the control performance of a TLL-NN architecture 
with 400 local linear region. While there are a number of techniques that can 
be used to train the resulting NN, for simplicity, we utilize Imitation 
learning where the NN is trained in a supervised fashion from data collected 
from an expert controller. In particular, we designed an expert controller that 
stabilizes the inverted pendulum; we used Pessoa 
\cite{MazoPESSOAToolEmbedded2010} to design our expert using the parameter 
values specified above. In particular, we tasked Pessoa to design a 
zero-order-hold controller that stabilizes the inverted pendulum in a subset 
$X_\text{spec} = [-1, 1] \negthinspace \times \negthinspace [-0.5, 0.5]$: i.e. 
the controller should transfer the state of the system to this specified set 
and keep it there for all time thereafter. 
From this expert controller, we collected 8400 data points of state-action 
pairs; this data was obtained by uniformly sampling the state space. We then 
used Keras \cite{chollet2015keras} to train the TLL NN 
using this data. 
%
Finally, we simulated the motion of the inverted pendulum using this TLL NN 
controller. Shown in \cref{fig:x0p7_v0p5} and \cref{fig:xm0p4_v1} are the state 
and control trajectories for this controller starting from initial state $[0.7 
,0.5]$ and $[-0.4,1]$, respectively. In both, the TLL controller met the same 
specification 
used to design the expert. 


\vspace{-2mm}

\bibliographystyle{ieeetr} %
\bibliography{mybib} %


\clearpage %
\begin{proofs} %
	\appendix


\section{Proofs} 
\label{sec:proofs}

\subsection{Proof of \cref{prop:eta_proposition}} 
\label{sub:proof_of_prop:eta_proposition}
\begin{proof}
	Choose $\eta = \eta(\mu) \leq \frac{\mu}{2 \cdot 3 \cdot K_\text{cont}}$ 
	and use Lipschitz continuity of $\Psi$.
\end{proof}

\subsection{Proof of \cref{prop:tau_proposition}} 
\label{sub:proof_of_prop:tau_proposition}
\begin{proof}
	Let $\mathcal{K}$ be the bound on $f$ as stated in 
	\cref{def:control_system}. Then by \eqref{eq:feedback_integral_eq} we have
	\begin{align}
		\lVert x - \zeta_{x\Upsilon}(t) \rVert &= \left\lVert \int_0^t f(\zeta_{x\Upsilon}(\sigma), \Upsilon(\zeta_{x\Upsilon}(\sigma) )) d\sigma \right\rVert \\
		&\leq  \int_0^t \left\lVert 
		f(\zeta_{x\Upsilon}(\sigma), \Upsilon(\zeta_{x\Upsilon}(\sigma) ))
		\right\rVert
		d\sigma  \\
		&\leq \int_0^t \mathcal{K} d\sigma = \mathcal{K} t.
	\end{align}
	Hence, choose $\tau = \tau(\mu) \leq \frac{\eta(\mu)}{\mathcal{K}}$ and the 
	conclusion follows.
\end{proof}

\subsection{Proof of \cref{prop:closed_loop_control_bounds}} 
\label{sub:proof_of_prop:closed_loop_control_bounds}
\begin{proof}
By the triangle inequality, we have:
	\begin{align}
		&\lVert 
				\Upsilon(\zeta_{x\Upsilon}(t)) - 
				\Psi(\zeta_{x\Psi}(t))
		\rVert \notag\\
		&\quad= \lVert 
				\Upsilon(\zeta_{x\Upsilon}(t)) - 
				\Psi(\zeta_{x\Upsilon}(t)) + 
				\Psi(\zeta_{x\Upsilon}(t)) -
				\Psi(\zeta_{x\Psi}(t))
		\rVert \notag\\
		&\quad=
		\lVert 
				\Upsilon(\zeta_{x\Upsilon}(t)) - 
				\Psi^\prime(\zeta_{x\Upsilon}(t)) + \notag \\
		&\qquad\qquad  \Psi(\zeta_{x\Upsilon}(t)) -
				\Psi(x) +
				\Psi(x) -
				\Psi(\zeta_{x\Psi}(t))
		\rVert \notag\\
		&\quad \leq
		\lVert 
				\Upsilon(\zeta_{x\Upsilon}(t)) - 
				\Psi(\zeta_{x\Upsilon}(t))
		\rVert +
		\lVert\Psi(\zeta_{x\Upsilon}(t)) -
				\Psi(x) \rVert + \notag\\
		&\qquad\qquad 
		\lVert
				\Psi(x) -
				\Psi(\zeta_{x\Psi}(t))
		\rVert.
		\label{eq:prop3_inequality}
	\end{align}
The first term in \eqref{eq:prop3_inequality} is bounded by $\lVert \Upsilon - 
\Psi \rVert_\infty$. Now consider the second term. By 
\cref{prop:tau_proposition}, $\lVert \zeta_{x\Upsilon}(t) - x \rVert \leq 
\eta$; thus, by \cref{prop:eta_proposition} we conclude that 
$\lVert\Psi(\zeta_{x\Upsilon}(t)) - \Psi(x) \rVert \leq \mu/3$. The final term 
is likewise bounded by $\mu/3$ for the same reasons. Thus, the conclusion 
follows.
\end{proof}

\subsection{Proof of \cref{prop:gronwall_lemma}} 
\label{sub:proof_of_prop:gronwall_lemma}
\begin{proof}
	By definition and the properties of the integral, we have:
	\begin{align}
		&\lVert \zeta_{x\Upsilon}(t) - \zeta_{x\Psi}(t) \rVert \notag \\
		&=
		\lVert
		\int_0^t 
			f(\zeta_{x\Upsilon}(\sigma),\Upsilon(\zeta_{x\Upsilon}(\sigma)))
			-
			f(\zeta_{x\Psi}(t), \Psi(\zeta_{x\Psi}(t)))
			d\sigma
		\rVert \notag \\
		&\leq
		\int_0^t 
			\lVert
			f(\zeta_{x\Upsilon}(\sigma),\Upsilon(\zeta_{x\Upsilon}(\sigma)))
			-
			f(\zeta_{x\Psi}(t), \Psi^\prime(\zeta_{x\Psi}(t)))
			\rVert
			d\sigma \notag \\
		&\leq
		\int_0^t 
			K_x 
			\lVert
				\zeta_{x\Upsilon}(\sigma)
				-
				\zeta_{x\Psi}(\sigma)
			\rVert \notag \\
		&\qquad\qquad+
			K_u
			\lVert
				\Upsilon(\zeta_{x\Upsilon}(\sigma))
				-
				\Psi(\zeta_{x\Psi}(\sigma))
			\rVert
			d\sigma \notag \\
		&\leq
		\int_0^t 
			K_x 
			\lVert
				\zeta_{x\Upsilon}(\sigma)
				-
				\zeta_{x\Psi}(\sigma)
			\rVert
			+
			K_u
			\cdot \kappa ~
			d\sigma.
	\end{align}
	The claimed bound now follows directly from the Gr\"onwall Inequality 
	\cite{KhalilNonlinearSystems2001}.
\end{proof}

\subsection{Proof of \cref{thm:main_theorem}} 
\label{sub:proof_of_thm:main_theorem}

\begin{proof}
	By \cref{lem:cpwa_construction}, there is a CPWA, $\Upsilon\subcpwa$ that meets 
	the assumptions of \cref{lem:final_spec}, and whose number of linear 
	regions is upper-bounded by the quantity in \eqref{eq:region_count}. Thus, 
	we are done if we can find a TLL NN architecture to implement the CPWA 
	$\Upsilon\subcpwa$. But by \cite[Theorem 2]{FerlezAReNAssuredReLU2020}, just 
	such an architecture can be specified directly by the number of linear 
	regions needed (as in \eqref{eq:region_count}). This completes the proof.
\end{proof}


\end{proofs}

\appendix

\end{document}